\documentclass{article}
\usepackage{multirow,spconf,amsmath,graphicx,url,times,booktabs,tabularx}
\usepackage{amsthm,amssymb,mathrsfs}

\title{AUDIO-VISUAL SCENE CLASSIFICATION USING A TRANSFER LEARNING BASED JOINT OPTIMIZATION STRATEGY}
\name{Chengxin Chen$^{1,2}$ \qquad Meng Wang$^{1,2}$ \qquad Pengyuan Zhang$^{*1,2}$\thanks{This work is partially supported by the National Natural Science Foundation of China (No. 62071461).}\thanks{* Pengyuan Zhang is the corresponding author.}}
\address{$^1$Key Laboratory of Speech Acoustics and Content Understanding, Institute of Acoustics, CAS, China\\
$^2$University of Chinese Academy of Sciences, China\\
\{chenchengxin, wangmeng, zhangpengyuan\}@hccl.ioa.ac.cn}

\begin{document}
%
\maketitle
\begin{abstract}
Recently, audio-visual scene classification (AVSC) has attracted increasing attention from multidisciplinary communities. Previous studies tended to adopt a pipeline training strategy, which uses well-trained visual and acoustic encoders to extract high-level representations (embeddings) first, then utilizes them to train the audio-visual classifier. In this way, the extracted embeddings are well suited for uni-modal classifiers, but not necessarily suited for multi-modal ones. In this paper, we propose a joint training framework, using the acoustic features and raw images directly as inputs for the AVSC task. Specifically, we retrieve the bottom layers of pre-trained image models as visual encoder, and jointly optimize the scene classifier and 1D-CNN based acoustic encoder during training. We evaluate the approach on the development dataset of TAU Urban Audio-Visual Scenes 2021. The experimental results show that our proposed approach achieves significant improvement over the conventional pipeline training strategy. Moreover, our best single system outperforms previous state-of-the-art methods, yielding a log loss of 0.1517 and accuracy of 94.59\% on the official test fold.
\end{abstract}
\section{INTRODUCTION}
\label{sec:introduction}
Automatic scene classification aims to classify the input signal into one of several predefined scene labels and help machines understand the environments around them. It has enormous potential in the applications of human-computer interaction, intelligent robotics, smart video surveillance and autonomous driving~\cite{Xie2020}. However, enabling devices to accurately recognize the environment is a challenging task due to the noise disturbance and ambiguity of input signal. 

In the past decades, a variety of algorithms have been proposed for Acoustic Scene Classification (ASC)~\cite{Barchiesi2015} and Visual Scene Classification (VSC)~\cite{Wei2016} task. State-of-the-art solutions for ASC are based on spectral features, most commonly the Mel-scale filter bank coefficients (FBank), and convolutional neural network (CNN) architectures~\cite{Sakashita2018}\cite{Chen2019}\cite{Suh2020}. 
In addition, long-term window depicts the scene information with various time scales, resulting in a more discriminative feature~\cite{Chen2021}. 
By contrast, VSC has a longer history and more types of approaches, e.g.,  global attribute descriptors~\cite{Oliva2001}, learning spatial layout patterns~\cite{Jiang2012} and discriminative region detection~\cite{Zuo2014}. 
Moreover, using powerful deep models pre-trained on large-scale image datasets as feature extractor has demonstrated promising performance~\cite{Damodaran2019}.

In recent years, researchers have shown increasing interests on audio-visual scene classification (AVSC), which utilizes acoustic and visual information simultaneously to further improve the performance of scene classification\cite{Wang2021audiovisual}.
Prior works on the AVSC task mainly focused on training state-of-the-art models for ASC and VSC individually, then retrieving the intermediate embeddings of each model to train the scene classifier~\cite{Parekh2020}\cite{Wang2021b}\cite{Wang2021}\cite{Okazaki2021}. We name this two-stage procedure “pipeline training strategy”. In this way, the extracted embeddings contain abundant information within each modality, and speed up the convergence of the classifier. 
However, the complementarity and redundancy across modalities are neglected.
To model the interactions within and across modalities simultaneously, various fusion methods have been proposed on other multi-modal fields\cite{Huang2020}\cite{Gao2019}\cite{Xu2019}. 
Most of these methods perform well on time-sequential related tasks using recurrent neural network (RNN). However, the AVSC task is less sensitive to the sequential relations. 

In this paper, we propose a “joint training strategy” for the AVSC task to model the interactions within and across modalities in a unified framework. 
The main contribution of this work is threefold.
First, we introduce long-term scalogram to the AVSC task inspired by~\cite{Chen2021}, and explore a frame-level alignment for audio-visual streams at the front end.
Second, we retrieve the bottom layers of pre-trained image models to serve as deep visual representation extractor to overcome the problem of image data sparsity.
Third, we propose to jointly optimize the acoustic encoder and scene classifier. 
In this way, the acoustic encoder is able to interact with visual representations to construct more discriminative audio-visual embeddings for scene classifier.
Additionally, it is the further research based on our prior work~\cite{Wang2021b}, which ranked first place in Task 1B of the challenge of Detection and Classification of Acoustic Scenes and Events (DCASE) in 2021.


\section{METHODOLOGY}
\label{sec:methodology}
\subsection{Model Structure}
\label{ssec:model}
\begin{figure}[htb]
\centering
\includegraphics[width=8.cm]{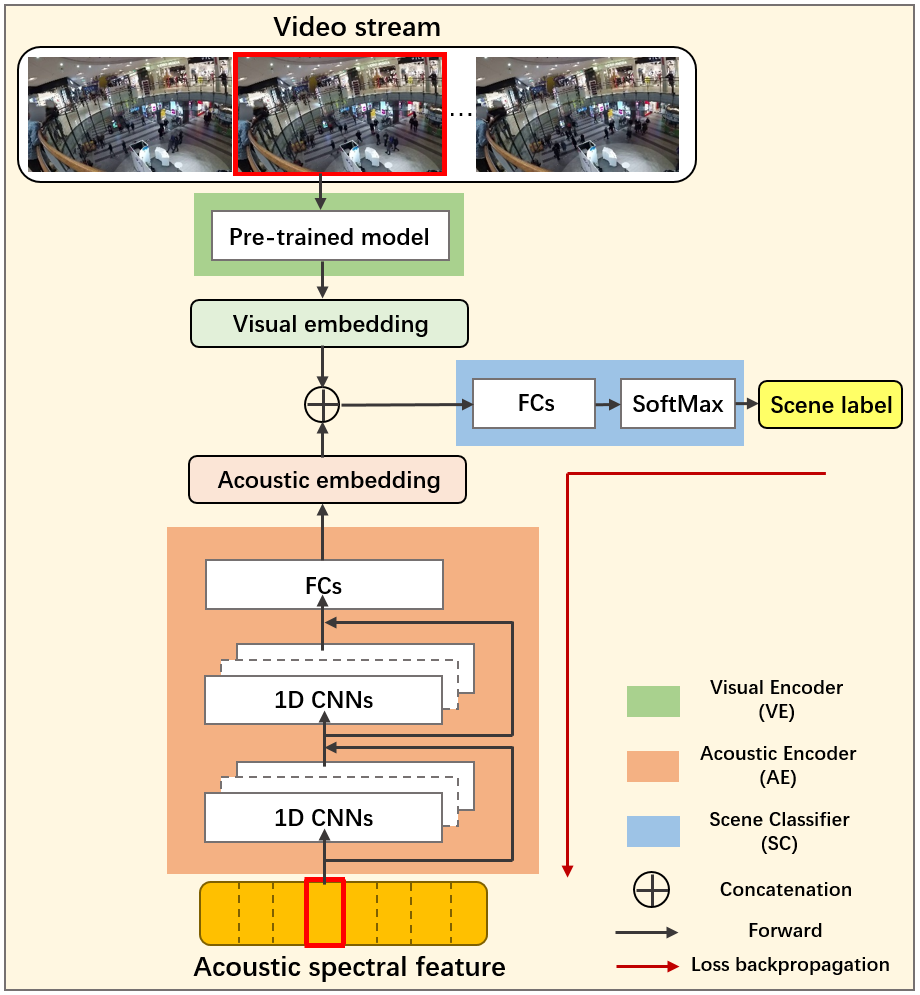}
\caption{Block diagram of the proposed framework.}
\label{fig:overview}
\end{figure}

Fig.~\ref{fig:overview} illustrates our proposed joint optimization framework for the AVSC task. The system is built with three modules: the acoustic encoder (AE), the visual encoder (VE) and the scene classification module (SC). 
The bottom layers of pre-trained image model are utilized as VE to extract deep representations, and the parameters are frozen during training. 
The structure of AE is based on 1D-DCNN~\cite{Chen2021}. 
We introduce residual learning~\cite{He2016} and name it “res-DCNN”. Specifically, we use a convolutional layer to extract features and an extra convolutional layer to match the dimension of input and output of each CNN block.
SC is built with stacked fully connected layers followd by a SoftMax layer, and the parameters of it are updated along with AE during training.

\subsection{Transfer Learning Based Joint Optimization}
\label{ssec:transfer}
\subsubsection{Pre-processing of Audio-visual Stream}
\label{sssec:prepro}

The audio files in the development dataset~\cite{Wang2021dataset} are recorded in binaural way. To extract features, the magnitude of the STFT spectrum is first calculated every 171 {\it ms} over 512 {\it ms} windows on the raw signal resampled at 16 kHz. Then the wavelet filters are applied directly to the STFT spectrum to obtain scalogram feature. 
For more details of the wavelet filters, please refer to~\cite{Chen2021}. We also extract long-term FBank for comparison. The only difference is that we apply Mel-scale filters to the STFT spectrum. 
For all features, we use average and difference channel instead of left and right channel.

On the video stream, we perform down-sampling with a frame rate of 1 fps since the images in the same video segment vary little. Then the acoustic frames and visual frames of the same video segment are time-aligned to construct our training dataset 
$ \mathcal{D} = \{(a_{1},v_{1},s_{1}),...,(a_{N},v_{N},s_{N})\}$, 
where $a_{i}, v_{i}, s_{i}$ denote the $i$th acoustic frame, visual frame and scene label, respectively.

\subsubsection{Pre-training of Visual Encoder}
\label{sssec:pretrain}
In this stage, we utilize the powerful image models developed in the computer vision area to obtain robust representations of visual frames. 
To validate the generality of our approach, we adopt the recently released EfficientNetV2-Small~\cite{Tan2021} and the classic ResNet50~\cite{He2016}. 
After downloading the open-source models pre-trained on ImageNet~\cite{Deng2009}, we fine-tune them on Places365~\cite{Zhou2018}, a large image dataset for scene recognition, to adjust the models to be more suitable for scene classification. 
Different from the conventional pipeline training strategy, we do not extract visual embeddings in advance.

\subsubsection{Joint Training of Acoustic Encoder and Scene Classifier}
\label{sssec:e2e train}
Given a training sample $(a_{i}, v_{i}, s_{i}) \in \mathcal{D}$, the intermediate representation $e_{i}$ can be calculated by
\begin{equation}
 e_{i} = {\rm AE}(a_{i}) \oplus {\rm VE}(v_{i}),
\end{equation}
where $\oplus$ denotes concatenation operation. The cross-entropy loss of SC’s output, denoted as $L_{scene}$, is defined by
\begin{equation}
L_{scene} = \sum_{i=1}^{N} –log P(s_{i} | {\rm SC}(e_{i})).
\end{equation}
The performance of SC is optimized by minimizing $L_{scene}$. We propose to update the parameters of AE and SC simultaneously through gradient decent, while the parameters of VE are frozen. 
In this way, VE is initialized with the ability to extract the global scene attribute of images, while AE learns to cooperate with VE to generate a more discriminative and compact representation. Moreover, data augmentation methods could be applied on raw inputs during training procedure.

We have tried to randomly initialize and train the whole model including VE in a totally end-to-end way. However, it took much more time for the system to converge and the final results were not satisfactory. To a certain extent, our approach is a trade-off between the conventional pipeline and end-to-end training strategy. 

\section{EXPERIMENTAL SETUP}
\label{sec:experiments}

\subsection{Dataset}
\label{ssec:dataset}
The development dataset of TAU Urban Audio-Visual Scenes 2021~\cite{Wang2021dataset} contains 34 hours of synchronized audio-visual data from 10 European cities, provided in files with a length of 10 seconds. It consists of 10 scene classes, including airport, metro station, public square, etc. The official training and test fold consist of 8646 and 3645 files, respectively. In our experiments, approximately 10\% of the training fold was randomly selected and reserved as the validation fold for hyperparameter fine-tuning and early stopping.

\subsection{Implementation Details}
\label{ssec:implement}
The detailed hyperparameters of the model are shown in Table~\ref{tab:details}. 
The notation ``2-3 Conv(pad=0,stride=1)-4-BN-ReLU-AvgPooling(pad=1,stride=2)'' denotes a convolutional kernel with 2 input channels, 4 output channels and a size of 3, followed by batch normalization and ReLU activation, finally an average pooling layer. 
We only used image data augmentation methods, including RandomResizedCrop, RandomHorizontalFlip, ColorJitter~\cite{paszke2019pytorch} and their combinations.
Besides, we used stochastic gradient descent (SGD) with cosine annealing and warm restart~\cite {Loshchilov2017} to optimize the model, with a batch size of 256 and max epochs of 150. 
The maximum and minimum learning rates were set to 1e-2 and 1e-5, respectively. The models with best validation loss were retained. 

\begin{table}[htbp]
\centering
\caption{\label{tab:details} Details of model structure and configuration.}
\scriptsize
\begin{tabular}{cl}
\toprule
Model                  & Settings                                                        \\
\midrule
\multirow{2}{*}{Input} & Frame size of scalogram: 2*290; FBank: 2*256      \\
                       & Image size for EffNetV2-S: 3*288*288; ResNet50: 3*224*224 \\
\midrule                       
\multirow{7}{*}{AE}    & 2-3 Conv(pad=0,stride=1)-4-BN-ReLU-AvgPooling(pad=1,stride=2)   \\
                       & 4-3 Conv(pad=0,stride=1)-8-BN-ReLU-AvgPooling(pad=1,stride=2)   \\
                       & 8-3 Conv(pad=0,stride=1)-16-BN-ReLU-AvgPooling(pad=1,stride=2)  \\
                       & 16-3 Conv(pad=0,stride=1)-32-BN-ReLU-AvgPooling(pad=1,stride=2) \\
                       & Flatten and concatenate input as well as Conv's output          \\
                       & Linear(2048 units)-BN-ReLU-Dropout(p=0.5)                       \\
                       & Linear(1024 units)-BN-ReLU                                      \\
\midrule                       
VE                     & Refer to~\cite{He2016} and~\cite{Tan2021}, and remove the final classification layer  \\
\midrule
\multirow{2}{*}{SC}    & Linear (1024 units)-BN-ReLU-Dropout(p=0.5)                      \\
                       & Linear (10 units)-SoftMax                                               \\
\bottomrule                       
\end{tabular}
\end{table}
In the test stage, we followed the official setup and split the test fold into 1-second segments, with a total number of 36450. Since the model was trained on frame-level, we took the average of the frame-wise probability distribution in the same segment as the final output. 

\subsection{Baseline Systems}
\label{ssec:base}

Three baseline systems were constructed to make a comparison with our proposed joint optimization audio-visual system.\\
\textbf{Audio-only system:} The model consists of two modules, AE and SC, as depicted in Fig.~\ref{fig:overview}. Only the audio files of the dataset were used to evaluate the system.\\
\textbf{Video-only system:} The model consists of two modules, VE and SC, as depicted in Fig.~\ref{fig:overview}. To make a fair comparison, VE was also pre-trained as described in Sec 2.2 and the parameters of it were frozen during training. Only the images of the dataset were used to evaluate the system.\\
\textbf{Pipeline audio-visual system:} First, we retrieved the trained AE of the above audio-only system and the pre-trained VE to extract acoustic and visual embeddings, respectively. The embeddings were then used to train SC.

\section{RESULTS AND DISCUSSION}
\label{sec:results}

\subsection{Experimental Results}
\label{ssec:exp_results}
Evaluation of systems was performed using two metrics as suggested in~\cite{Wang2021audiovisual}: multi-class cross-entropy (log-loss) and accuracy. Both metrics are calculated as average of the class-wise performance, and the log-loss was the principle metric of DCASE2021 task 1B.

\begin{table}[htbp]
\centering
\caption{\label{tab:evaluation of uni-modal systems}Evaluation of different uni-modal systems.}
\footnotesize
\begin{tabular}{ccccc}
\toprule
System                         & Input   & Backbone      & Log-loss        & Acc/\%    \\ 
\midrule
\multirow{2}{*}{Audio-only}      & FBank     & Res-DCNN  & 0.6968          & 76.08          \\  
                            & scalogram & Res-DCNN  & \textbf{0.6325} & \textbf{77.98} \\ 
\midrule
\multirow{2}{*}{Video-only}      & raw image & ResNet50   & 0.3939          & 87.03          \\ 
                            & raw image & EffNetV2-S & \textbf{0.2731} & \textbf{90.91} \\ 
\bottomrule
\end{tabular}
\end{table}

As shown in Table~\ref{tab:evaluation of uni-modal systems}, in the case of audio-only system, using long-term scalogram as acoustic feature achieved a lower log-loss and higher accuracy compared with FBank. For video-only system, the EfficientNetV2-Small backbone demonstrated more strength to extract high-level representations from raw images. 

\begin{table}[htbp]
\centering
\caption{\label{tab:evaluation of audio-visual systems}Evaluation of different audio-visual systems.}
\footnotesize
\begin{tabular}{ccccc}
\toprule
Training strategy              & Feature(A) & Model(V)   & Log-loss        & Acc/\%         \\ 
\midrule
\multirow{4}{*}{Pipeline} & FBank      & ResNet50   & 0.3808          & 88.87          \\  
                          & FBank      & EffNetV2-S & 0.2146          & 92.84  \\  
                          & scalogram  & ResNet50   & 0.3506          & 89.29          \\  
                          & scalogram  & EffNetV2-S & \textbf{0.2055} & \textbf{93.14} \\ 
\midrule
\multirow{4}{*}{Joint}  & FBank      & ResNet50   & 0.2664          & 91.13          \\
                          & FBank      & EffNetV2-S & 0.1788          & 93.45          \\
                          & scalogram  & ResNet50   & 0.2495          & 91.55          \\
                          & scalogram  & EffNetV2-S & \textbf{0.1517} & \textbf{94.59} \\
\bottomrule
\end{tabular}
\end{table}

In Table~\ref{tab:evaluation of audio-visual systems}, we summarize the performance of different combination of acoustic features and pre-trained image models. 
Compared with pipeline training strategy, we can observe a consistent performance gain of all the combinations when using joint training strategy. 
Besides, the superiority of acoustic features and pre-trained image models could further accumulate.
Finally, the best single system turned out to be the joint optimization audio-visual system using long-term scalogram as acoustic feature and EfficientNetV2-Small as pre-trained image model. 
The following experiments would all be performed on this combination.

\subsection{Ablation Study}
\label{ssec:ablation}
The key point of the joint training strategy lies in two aspects: (1) AE is trainable in the joint training. (2) Raw images are used as inputs in the joint training, and data augmentation methods are applied in the training procedure.
Table~\ref{tab:ablation result} presents the results of evaluation on the key factors. \uppercase\expandafter{\romannumeral1} and \uppercase\expandafter{\romannumeral4} represent the pipeline and joint training strategy, respectively. Comparing \uppercase\expandafter{\romannumeral1} and \uppercase\expandafter{\romannumeral2}, introducing joint optimization without data augmentation lead to performance degradation. The reason may be that the extracted embeddings were not diverse enough to train a robust system. Comparing \uppercase\expandafter{\romannumeral1} and \uppercase\expandafter{\romannumeral3}, using augmented raw images as input could significantly improve the performance. On the basis of it, the introducing of trainable AE could further boost the performance of system comparing \uppercase\expandafter{\romannumeral3} and \uppercase\expandafter{\romannumeral4}.

\begin{table}[htbp]
\centering
\caption{\label{tab:ablation result}Abation study on the audio-visual systems.}
\small
\begin{tabular}{ccccc}
\toprule
No. & Feature(V) & Model(A)    & Log-loss & Acc/\% \\
\midrule
\uppercase\expandafter{\romannumeral1} & embedding  & pre-trained & 0.2055   & 93.14   \\
\uppercase\expandafter{\romannumeral2} & embedding  & trainable   & 0.2144   & 92.39   \\
\uppercase\expandafter{\romannumeral3} & raw image   & pre-trained & 0.1591   & 94.31   \\
\uppercase\expandafter{\romannumeral4} & raw image   & trainable   & \textbf{0.1517}   & \textbf{94.59}   \\
\bottomrule
\end{tabular}
\end{table}

\subsection{Visualization Analysis}
\label{ssec:visualize}
\begin{figure}[htb]
\centering
\scalebox{0.9}{
\includegraphics[width=8.cm]{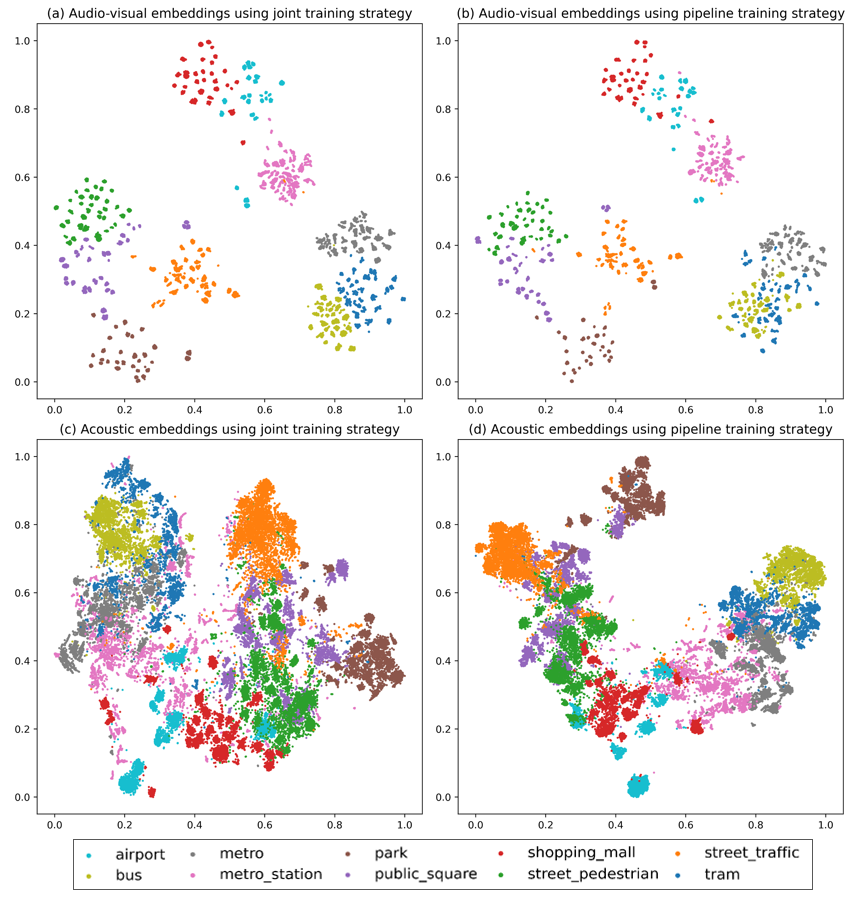}}
\caption{The t-SNE visualization of embedding distributions using different training strategies.}
\label{fig:tsne}
\end{figure}

We plot the t-SNE~\cite{van2008visualizing} to visualize the embeddings of the test fold using different training strategies. As shown in Fig.~\ref{fig:tsne} (a) and (b), in the pipeline setting, the points of the metro, bus, and tram are entangled with each other. By contrast, we can observe clear boundaries on the clusters of these categories in the joint optimization setting. 
However, comparing Fig.~\ref{fig:tsne} (c) and (d), the acoustic embeddings of different categories using joint training strategy are more dispersed, which indicates that the optimization direction of acoustic encoder have turned into constructing more discriminative audio-visual embeddings instead of better acoustic embeddings.

\subsection{Comparison With Prior Works}
\label{ssec:comparison}
In Table~\ref{tab:comparison}, we compare the performance of our proposed joint optimization framework with the state-of-the-art methods of DCASE2021 task 1B. For fair comparison, we select the best audio-visual single systems from the technical reports~\cite{Wang2021b}~\cite{Wang2021}~\cite{Yang2021}. Our approach achieves the lowest log-loss of 0.152 and the highest accuracy of 94.6\%. 
All the other systems, including official baseline, adopted the pipeline training strategy. 
\cite{Wang2021} and~\cite{Yang2021} explored various state-of-the-art audio and visual models pre-trained on large scale datasets, and applied lots of data augmentation methods on both acoustic features and images.
The key difference of our approach is that we aim to directly derive the best audio-visual embeddings,  instead of obtaining the best acoustic and visual embeddings separately.

\begin{table}[htbp]
\centering
\caption{\label{tab:comparison}Comparison with previous state-of-the-art methods.}
\small
\begin{tabular}{cccc}
\toprule
Method            & Training strategy & Log-loss & Acc/\% \\
\midrule
Official baseline~\cite{Wang2021dataset} & Pipeline          & 0.658    & 77.0   \\
\midrule
Wang et al.~\cite{Wang2021b}      & Pipeline          & 0.159    & 94.1   \\
Wang et al.~\cite{Wang2021}       & Pipeline          & 0.183    & 93.8   \\
Yang et al.~\cite{Yang2021}      & Pipeline          & 0.223    & 93.9   \\
\midrule
Our proposed      & Joint         & \textbf{0.152}   & \textbf{94.6}   \\
\bottomrule
\end{tabular}
\end{table}

\section{CONCLUSION}
\label{sec:conclusion}

In this paper, we propose a joint training strategy for audio-visual scene classification. Compared with the conventional pipeline setting, our approach can derive more discriminative audio-visual embeddings. Besides, using the long-term scalogram as acoustic feature demontrates better performance than FBank in all of our experiments. Results of ablation study show that the data augmentation methods applied on the raw images play a vital role on the joint training procedure. 
In the future, we intend to explore the knowledge distillation methods across modalities to tackle the modality asynchronous or missing problems in the real world.



\vfill\pagebreak

\bibliographystyle{IEEEbib.bst}
\footnotesize
\bibliography{refs.bib}

\end{document}